\documentclass[10pt,twocolumn,a4paper]{elsarticle}
\usepackage[T1]{fontenc}
\usepackage{times}
\usepackage{epsfig}
\usepackage{graphicx}
\usepackage{amsmath}
\usepackage{amssymb}
\usepackage{booktabs}
\usepackage{tabulary}
\usepackage[dvipsnames, svgnames]{xcolor}
\usepackage[caption=false]{subfig}
\usepackage[british,american]{babel}
\usepackage{soul}
\usepackage{xspace}
\usepackage{adjustbox}
\usepackage{array}
\usepackage{dblfloatfix}
\usepackage{transparent}
\usepackage{algorithm}
\usepackage{listings}
\usepackage{bbold}
\usepackage{arydshln}
\usepackage{url}
\usepackage[switch]{lineno}
\usepackage{textpos}
\usepackage{multicol}
\usepackage{geometry}
\usepackage{hyperref}
\usepackage{float}
\usepackage{xpatch}
\usepackage{makecell}
\usepackage{ulem}
\usepackage{etoolbox}
\usepackage{lipsum}
\usepackage{everyhook}

\journal{Elsevier}

\biboptions{authoryear}

\makeatletter
\DeclareRobustCommand\onedot{\futurelet\@let@token\@onedot}
\def\@onedot{\ifx\@let@token.\else.\null\fi\xspace}
\def\ie{\emph{i.e}\onedot}
\def\eg{\emph{e.g}\onedot}

\usepackage{etoolbox}
\makeatletter
\AfterEndEnvironment{algorithm}{\let\@algcomment\relax}
\AtEndEnvironment{algorithm}{\kern2pt\hrule\relax\vskip3pt\@algcomment}
\let\@algcomment\relax
\newcommand\algcomment[1]{\def\@algcomment{\footnotesize#1}}
\renewcommand\fs@ruled{\def\@fs@cfont{\bfseries}\let\@fs@capt\floatc@ruled
  \def\@fs@pre{\hrule height.8pt depth0pt \kern2pt}%
  \def\@fs@post{}%
  \def\@fs@mid{\kern2pt\hrule\kern2pt}%
  \let\@fs@iftopcapt\iftrue}
\makeatother

\definecolor{citecolor}{HTML}{0071bc}
\definecolor{HighlightImprove}{HTML}{000000}
\definecolor{HighlightDecline}{HTML}{000000}
\definecolor{greyColor}{HTML}{424242}

\newcommand{\app}{\raise.17ex\hbox{$\scriptstyle\sim$}}

\newcolumntype{x}[1]{>{\centering\arraybackslash}p{#1pt}}
\newcolumntype{y}[1]{>{\raggedright\arraybackslash}p{#1pt}}
\newcolumntype{z}[1]{>{\raggedleft\arraybackslash}p{#1pt}}
\newlength\savewidth
\newcommand{\tablestyle}[2]{\setlength{\tabcolsep}{#1}\renewcommand{\arraystretch}{#2}\centering\footnotesize}

\makeatletter\renewcommand\paragraph{\@startsection{paragraph}{4}{\z@}
  {.5em \@plus1ex \@minus.2ex}{-.5em}{\normalfont\normalsize\bfseries}}\makeatother

\newcommand\blfootnote[1]{%
  \begingroup
  \renewcommand\thefootnote{}\footnote{#1}%
  \addtocounter{footnote}{-1}%
  \endgroup
}

\begin{document}
\onecolumn

\title{Differentiable $N$-gram Objective on Abstractive Summarization}
\author{
  Yunqi Zhu\textsuperscript{a,b}, \sep
  Xuebing Yang\textsuperscript{b,*}, \sep
  Yuanyuan Wu\textsuperscript{a}, \sep
  Mingjin Zhu\textsuperscript{c}, \sep
  Wensheng Zhang\textsuperscript{a,b,*}
  \\ \vspace{1em}
  \emph{\textsuperscript{a}School of Information and Communication Engineering, Hainan University, Haikou, China}
  \\
  \emph{\textsuperscript{b}Research Center of Precision Sensing and Control, Institute of Automation, Chinese Academy of Sciences, Beijing, China}
  \\
  \emph{\textsuperscript{c}Shien-Ming Wu School of Intelligent Engineering, South China University of Technology, Guangzhou, China}
}

\setlength{\columnsep}{2em}

\begin{abstract}
  ROUGE is a standard automatic evaluation metric based on $N$-gram for sequence-to-sequence tasks like abstractive summarization, while cross-entropy loss is an essential objective that optimizes at unigram level for neural network language models. In this paper 
  we present differentiable $N$-gram objectives, attempting to alleviate the discrepancy between training and evaluating criteria. The novelty of our work is the objective does not ceil the number of matched sub-sequences by the ground truth count of $N$-gram in reference sequence and weights the matched sub-sequences equally. Therefore, our proposed objective can maximize the probabilistic weight of matched sub-sequences. We jointly optimize cross-entropy loss and the objective, providing decent evaluation scores enhancement including ROUGE over abstractive summarization datasets CNN/DM and XSum, outperforming competitive $N$-gram objectives. 

\end{abstract}


\twocolumn[{%
  \begin{@twocolumnfalse}
    \maketitle
  \end{@twocolumnfalse}
}]

\section{Introduction}

Automatic text summarization has attracted much attention in recent years. 
Text summarization can be classified into extractive or abstractive summarization.
The goal of extractive summarization is identifying and extracting a few important and comprehensive sentences from the source docuements \citep{zhou-etal-2018-neural-document, Nallapati_Zhai_Zhou_2017}.
Abstractive summarization compresses the context of  source material and rewrites the text into shorter version while retaining the meanings \citep{ELKASSAS2021113679}.
Abstractive summarization is commonly viewed as a sequence-to-sequence (seq2seq) learning process \citep{Sutskever2014}.
We concentrate on abstractive approaches in this paper.
\blfootnote{
  {\small * Corresponding authors. 
  }}
\blfootnote{
 {\small  Email addresses: zhuyunqi96@163.com (Y. Zhu); yangxuebing2013@ia.ac.cn (X. Yang);  wyuanyuan82@163.com (Y. Wu); 
    zhumingjin19re@foxmail.com (M. Zhu); 
    zhangwenshengia@hotmail.com (W. Zhang) 
    }}
\blfootnote{
  {\small Code is available at: github.com/zhuyunqi96/ngramObj
}}

Seq2seq language models \citep{Bahdanau2015} with attention mechanism have dominated various natural language processing (NLP) downstream tasks\citep{rush2015, chopra2016, Vaswani2017, Radford2018, Devlin2019, Raffel2019, Lewis2020, Qi2020}. 
In recent years, large pre-trained language models with self-supervised autoencoder and autoregressive generation have greatly improved the semantic quality of summarization.
It is also worth noticing that language model with representation learning methods (\eg, bottom-up \citep{gehrmann2018bottomup}, contrastive learning \citep{Xu_Zhang_Wu_Wei_2022} and external faithful signals \citep{dou-etal-2021-gsum}) can effectively improve the quality of machine generated summaries.
Recent works investigated external commonsence enhanced representation \citep{yang-human-Like-2021} and topic-oriented semantic representation \citep{nguyen-etal-2021-enriching} for abstractive summarization.
To efficiently evaluate the consistency, conciseness and faithfulness of machine generated summaries, deep neural network based evaluation is receiving increasing attention \citep{kryscinski2020factCC, LabanSBH22Summac, liu-etal-2022-brio, ladhak-etal-2022-faithful}.

Most of the language models use cross-entropy loss as the learning objective, which is simple and standard for seq2seq model. 
Automatic evaluation methods based on $N$-gram, \textit{e.g.},  recall-oriented understudy for gisting evaluation (ROUGE) metric \citep{Lin2004} and bilingual evaluation understudy (BLEU) metric \citep{Papineni2002}, are widely-used for generation tasks such as text summarization \citep{Bhandari2020} and machine translation. Meanwhile the widely used cross-entropy objective merely considers exact token matching from the reference text by rewarding the probability of ground truth token and diminishing the others. 
The discrepancy between the learning objective and the evaluation objective (usually referred as {\emph{exposure bias}} \citep{Ranzato2016}) may make the language model miss some reliable alternative sub-sequences, while humanly text summarization tends be tolerant of alternative language expressions or shifted sub-sequences.

\citep{yang-etal2018} and \citep{li-piccardi-2021} used generative adversarial neural network to optimize BLEU metric on machine translation.
\citep{Ranzato2016} and \citep{Wu-etal2016} tried bridging the discrepancy by rewarding BLEU and grammaticality evaluation utility (GLEU) metrics for text generation with reinforcement learning \citep{Williams1989}.
\citep{paulus2018} applied reinforcement learning approach on text summarization. However, high variance of sampling can make the process of reinforcement learning unstable and less reproducible. \citep{ma2018bagofwords} and \citep{Shao2020BoN} proposed bag-of-words \citep{Joachims1998} and bag-of-ngrams (BoN) as sequence-level training target respectively. 
They intented to minimize the gap between the probability distribution of every token in the output sequence and a set of token or $N$-gram from reference sequence.
\citep{Zhukov2017} introduced a lower bound approximation of expected BLEU score for sequence generation.
\citep{Casas2018} proposed a differentiable BLEU objective with approximation of $N$-gram matches using Gumbel-softmax \citep{jang2017}.
\citep{yavuz2018} attempted to develop approximation objective for the longest common sub-sequence.
\citep{shao2018greedy} proposed a differentiable probabilistic $N$-gram count objective by maximizing the probabilistic count of matched $N$-gram. $N$-gram matching is verified with the \texttt{argmax} result of the final output of the language model.
However their objective will stop rewarding the probabilistic count of an $N$-gram if its expectation reaches the corresponding count in reference sequence.
Although cross-entropy is the most contributing learning objective,
probabilistic $N$-gram objectives can serve  for representation learning to shape the probabilistic distribution of the hidden state output of a language model.
Since the probabilistic $N$-gram is independently recognized from the final hidden state, removing the $N$-gram count ceiling could enable the language model to reward complete $N$-gram matches in the training stage.
This is particularly important to achieve satisfying weighted linear combination for $N$-gram matches when weight vectors of the probabilistic $N$-gram count cannot define the simplex (i.e., sum of the weight vectors exceed 1).

Inspired by probabilistic $N$-gram count objective \citep{shao2018greedy} and BoN objective \citep{Shao2020BoN}, we want the learning objective to be flexible for $N$-gram matching which is not limited by the reference $N$-gram count ceiling, 
and expect the co-occurrences of $N$-gram have equal weights in the objective, 
yet retaining the capability of alleviating the \emph{exposure bias}. In this paper, we propose two differentiable $N$-gram objectives: 
1) differentiable $N$-gram rewards: a position-related $N$-gram matches objective that maximizes the probabilistic of matched $N$-gram between the output sequence and reference sequence. The matched sub-sequence is counted only if it shares the uniform position in candidate text and reference text;
2) differentiable $N$-gram matches: a position-unrelated $N$-gram matches objective that maximizes the probabilistic of matched $N$-gram between the output sequence and reference sequence. The matched sub-sequence is not constrained by positions. 
Unlike \citep{shao2018greedy, Shao2020BoN}, our $N$-gram rewards objective and $N$-gram matches objective value the matched $N$-gram equally, and continue optimizing a matched $N$-gram even if it exceeds the count of this $N$-gram in reference text.
We fine-tune the pre-trained model BART-base on abstractive summarization task, jointly optimizing the proposed objective and cross-entropy loss.
Finally, experiment results show that our algorithm outperforms the probabilistic $N$-gram count objective and BoN objective with cross-entropy loss on text summarization.

Our contributions are listed as follows.
\begin{itemize}
  \item Our proposal, jointly optimizing cross-entropy loss and $N$-gram objective that equally weights $N$-gram matches and does not have a ceiling for $N$-gram matches, can deliver decent evaluation scores improvement on text summarization and surpass alternative $N$-gram objectives. 
  \item We conduct extensive experimental evaluation on the CNN/DailyMail dataset and the XSum dataset using ROUGE scores, 
        BERTscore, Word Mover's Distance, FactCCX and SummaC\textsubscript{CONV} as summarization metrics, 
        confirming the improvement made by our proposed objectives.
  
\end{itemize}

\section{Background and Related Work}

\paragraph{ROUGE-N Metric.} ROUGE-N is a widely used metric that measures the co-occurrence of $N$-gram between the reference sequence and candidate sequence. Usually the F1 score of ROUGE-1, ROUGE-1 and ROUGE-L will be reported (abbreviated as R-1, R-2, R-L respectively). 
R-1 and R-2 measures overlap of unigram and bigram respectively, while R-L evaluates the longest common sub-sequence. ROUGE-N is calculated as follows:

\begin{small}
  \begin{equation}
    \textrm{ROUGE-N} = \frac{\sum_{S\in\{Ref\}} \sum_{{gram_n}\in{S}} Count_{\rm match} ({gram_n})}{\sum_{S\in\{Ref\}} \sum_{{gram_n}\in{S}} Count ({gram_n})}
    \label{eq:ROUGE-N}
  \end{equation}
\end{small}

\paragraph{Cross-Entropy Loss.} 
Optimizing cross-entropy (CE) loss is a fundamental approach to minimize the negative log-likelihood of probability distribution over the one-hot representation of reference text.
Denote $x$ as an input text with sequence length $T$, 
and $y$ as the softmax output of final hidden state $\{H_{i}\}_{i=1}^{T}$  with vocabulary size  $D$ of a language model. 
We typically feed the matrix into log computation.
For a reference sequence $\hat{y}$ = ($\hat{y}_{1}$,...,$\hat{y}_{T}$),
CE loss is computed as follows:

\begin{equation}
{
\mathcal{L}_{CE}(y,\hat{y}) = -\sum_{t=1}^{T}\log\frac{p(\hat{y}_{t}|y_{t},x)}{\sum_{d=1}^{D}p(d|y_{t},x)} }
\label{eq:Cross-entropy}
\end{equation}

CE loss excepts machine generated sequence to match every token at every position over reference sequence, which maximizes the probability of matching tokens while minimizes the probability of the others. 
This can lead to a discrepancy between the learning process and the evaluation metric, and may limit the expressiveness of language model. 
Since evaluating sequence generation task with $N$-gram is a standard and feasible automatic approach, the nature of CE loss may drop practical variances during the training. 
In specific, CE loss may penalize $N$-gram sub-sequence that does not occur in the precise position, which reduces possible alternatives such as synonymic sentences, sub-sequence reordering, practical sub-sequence shifting, active and passive voices, and token-level synonyms.

\paragraph{Probabilistic $N$-gram Count Objective.}
\citep{shao2018greedy} proposed a sequence-level $N$-gram matching objective. During training, they used the \texttt{argmax} result of the final output of seq2seq model, to accumulate the probabilistic count of the position-unrelated matches of $N$-gram, providing significant improvement on machine translation datasets. They examined probabilistic objectives of $N$-gram, BLEU and GLUE. The result of probabilistic 2-gram count precision (P-P2) surpassed the others.

Maximizing the product of probabilistic unit of an $N$-gram is the optimization target.
Given source input $x$, model parameters $\theta$, output sequence $y$, reference sequence $\hat{y}$ and $N$-grams $g$ = ($g_{1}$...$g_{n}$), and use tilde to denote probabilistic variables, then the probabilistic count of $g$ is computed as:
\begin{equation}
\tilde{C}_{y}(g) = \sum_{t=0}^{T-n}\prod_{i=1}^{n}\mathbb{1}\{g_{i}=y_{t+i}\}\cdot p(y_{t+i}|y_{<t+i}, x, \theta)
\label{eq:prob ngram count}
\end{equation}
where $\mathbb{1}\{\cdot\}$ is an indicator function, return 1 if the condition is satisfied otherwise return 0. Then, compute matching count of $N$-gram:
\begin{equation}
\tilde{C}_{y}^{\hat{y}}(g) = \textrm{min}(\tilde{C}_{y}(g), C_{\hat{y}}(g))
\label{eq:matching ngram count}
\end{equation}
and probabilistic precision of $N$-gram:
\begin{equation}
\tilde{p}_{n} = \frac{\sum_{g\in y} \tilde{C}_{y}^{\hat{y}}(g)} {\sum_{g\in y} \tilde{C}_{y}(g)}
\label{eq:prob precision $N$-gram}
\end{equation}

Finally, the objective of probabilistic $N$-gram count is:
\begin{equation}
  \mathcal{L}_{n}(\theta) = - \tilde{p}_{n}
  \label{eq:prob ngram count objective}
  \end{equation}

\paragraph{BoN Objective.}
\citep{Shao2020BoN} considered BoN objective to minimize the difference between the probability distribution of each generated tokens and the probability distribution of $N$-gram of reference sequence during training.
BoN objective can be jointly optimized with CE loss. 
For reference text $y$ with sequence length $T$ and token $t$, an $N$-gram $g$ = ($g_{1}$...$g_{n}$) of the reference text is defined as follows, 
\begin{equation}
\textrm{BoN}_{y}(g) = \sum_{t=0}^{T-n} \mathbb{1} \{ y_{t+1:t+n} = g \}
\label{eq:BoN-Y}
\end{equation}

Given a source sequence $x$, for seq2seq model with parameter $\theta$, the probability distribution of generated sequence $\textrm{BoN}_\theta$ is defined as :

\begin{equation}
\textrm{BoN}_{\theta}(g) = \sum_{t=0}^{T-n} \prod_{i=1}^{n} p(y_{t+i} = g_{i} | X, \theta)
\label{eq:BoN-theta}
\end{equation}

To minimize the difference between $\textrm{BoN}_\theta$ and $\textrm{BoN}_y$, BoN objective is:

\begin{equation}
\mathcal{L}_{BoN}(\theta) = \frac{ 2(T - n + 1) - \sum_{g} \textrm{min}(\textrm{BoN}_{\theta}(g), \textrm{BoN}_{y}(g)) } {2(T - n + 1)}
\label{eq:BoN-objective}
\end{equation}

\section{Method}

In this section, we will detail two differentiable $N$-gram objectives:
\textit{differentiable $N$-gram rewards} is a probabilistic positon-related $N$-gram matching objective;
\textit{differentiable $N$-gram matches} is a probabilistic positon-unrelated $N$-gram matching objective.
Finally we detail pre-trained language model BART-base implementation with proposed $N$-gram objectives.

\subsection{Differentiable $N$-gram rewards}
\label{sec:$N$-gram rewards method}

Inspired by probabilistic $N$-gram matching count \citep{shao2018greedy} and BoN \citep{Shao2020BoN}, the objective of differentiable $N$-gram rewards ($N$ $\geq$ 2) is maximizing the position-related matched $N$-gram between the reference summary and the candidate summary, where matched token is observed by checking maximum probabilities over the vocabulary through \texttt{argmax}. Note that
$N$-gram rewards objective does not seek to optimize the $N$-gram count difference, hence can also be jointly optimized with cross-entropy loss.

Unigram is excluded because the objective encourages probabilities of the token only when it is matched with the reference token. As unigram rewards can be considered as a limited case of cross-entropy loss, repeatedly and limitedly maximizing the probabilities with a different measure of objective could be redundant and is out of our consideration.

The $N$-gram rewards objective is computed as follows:

\begin{equation}
\mathcal{L}_{n \rm -gram\;rewards}(\theta) = 1 - \frac{\sum_{t=0}^{T-n} \mathit{Reward}(\cdot)} {T - n + 1}
\label{eq:$N$-gram rewards objective1}
\end{equation}

\begin{small}
  \begin{equation}
    \mathit{Reward}(\cdot)=\left\{
      \begin{aligned}
          & \frac{\prod_{i=1}^{n}  p(y_{t+i} = g_{i} | X)} {Count_{\rm match}g_{t\in\{cand\}}},\; g_{t\in\{cand\}}=g_{t\in\{ref\}}\\
          &\quad\quad\quad\quad\quad 0, \quad\quad\quad\quad g_{t\in\{cand\}} \neq g_{t\in\{ref\}}
      \end{aligned}
      \right.
  \label{eq:$N$-gram rewards objective2}
  \end{equation}
\end{small}

If $N$-gram $g_{t}\in\{cand\}$ is equivalent to $g_{t}\in\{ref\}$ (subscript $t$ represents they share the same index in the sequence), $\mathit{Reward}(\cdot)$ will return the product of probabilities for unigrams under $g$ over the count of this matched $N$-gram in the sequence, otherwise return 0.
The cumulative product represents the similarity between the matched $N$-gram and the ground truth $N$-gram.
The denominator in $\mathit{Reward}(\cdot)$ indicates that every matched $N$-gram weights equally. Furthermore, the objective offers possibility of exceeded matches of $N$-gram. It should be remarked that P-P2 \citep{shao2018greedy}, BoN \citep{Shao2020BoN} and our method all contain non-differentiable opertation in building the dictionary of $N$-grams from reference sequence.

Algorithm~\ref{alg:code reward} shows the PyTorch implementation of 2-gram rewards objective.
Let $x_{i}$ denote an input text. Given a mini-batch input $\mathbf{B}$ of size $b$, $\mathbf{B} = \{x_1...x_b\} $, 
we denote the final hidden state output of the language model for the batch as $lm\_logits = \{y_1...y_b\}$,
$y_i \in\mathbb{R}^{T \times D}$.
We map $lm\_logits$ with softmax function and slice $y_i$ with the actual text length of the corresponding input text (by removing \texttt{[PAD]} tokens), 
followed by the calculation of our proposed $N$-gram objective.

\subsection{Differentiable $N$-gram matches}
\label{sec:$N$-gram matches method}

Differentiable $N$-gram matches objective ($N$ $\geq$ 1) is a probabilistic position-unrelated matching objective, and the main difference from Section~\ref{sec:$N$-gram rewards method} appeares in Equation~\ref{eq:$N$-gram rewards objective2} by changing $\mathit{Reward}(\cdot)$ to $\mathit{Match}(\cdot)$. That is, 
$\mathit{Match}(\cdot)$ function remains same outputs as $\mathit{Reward}(\cdot)$, but the condition is updated:
\begin{equation}
  \mathcal{L}_{n \rm -gram\;matches}(\theta) = 1 - \frac{\sum_{t=0}^{T-n} \mathit{Match}(\cdot)} {T - n + 1}
  \label{eq:$N$-gram matches objective1}
\end{equation}
\begin{small}
  \begin{equation}
    \mathit{Match}(\cdot)=\left\{
      \begin{aligned}
          & \frac{\prod_{i=1}^{n}  p(y_{t+i} = g_{i} | X)} {Count_{\rm match}g_{t\in\{cand\}}},\; g_{t\in\{cand\}} \in g_{\in\{ref\}}\\
          &\quad\quad\quad\quad\quad 0, \quad\quad\quad\quad g_{t\in\{cand\}} \notin g_{\in\{ref\}}
      \end{aligned}
      \right.
  \label{eq:$N$-gram matches objective2}
  \end{equation}
\end{small}

The probabilistic $N$-gram matches are no longer restrained by positions, therefore the $N$-gram objective  basically corresponds to what ROUGE scores demand. 
As long as the $N$-gram is matched, differentiable probabilistic similarity of an $N$-gram is weighted.
Unigram is accepted this time because a matched unigram can be a position-unrelated matched token across output summary and reference summary.

Again, our objective does not ceil the probabilistic counts of $N$-gram. 
It proportionately values the contribution of multiple matches of an $N$-gram.
Algorithm~\ref{alg:code match} presents the PyTorch implementation of 2-gram matches objective,
where lines [17-35] realizes Equation~\ref{eq:$N$-gram matches objective1} and Equation~\ref{eq:$N$-gram matches objective2}.
Note that the input parameters are identical to Algorithm~\ref{alg:code reward}'s.

\vspace{-10pt}

\begin{algorithm}[t]
  \caption{\small{2-gram rewards in PyTorch-like style}}
  \label{alg:code reward}
  \algcomment{\fontsize{7pt}{0em}\selectfont 
  \textrm{all}: torch.all(), return True if all elements of the input are True.\\
  \textrm{max}: torch.max(), return maximum values of the elements and their indices.\\
  \textrm{prod}: torch.prod(), return the product of all elements in the vetor.
  }
  \definecolor{codeBLEU}{rgb}{0.25,0.5,0.5}
  \lstset{
    backgroundcolor=\color{white},
    basicstyle=\fontsize{7pt}{7pt}\rmfamily\selectfont,
    columns=fullflexible,
    breaklines=true,
    captionpos=b,
    commentstyle=\ttfamily\selectfont\fontsize{7pt}{7pt}\color{black},
    keywordstyle=\color{black}\fontsize{7pt}{7pt},
    numbers=left,
    numbersep=-8pt,
    emph={
      softmax_prob,
      lm_logits
      summary_prob,
      prob_values,
      cand_indices,
      summary_prob,
      refsum_i,
      vocab,
      seq_len,
      lm_logits,
      twoGram_loss_i,
      twoGrams,
      s,
      k,
      i,
      v,
      each_gram,
      num_grams,
      prob,
      b_i
    },
    emphstyle={\itshape},
    morecomment=[l]{//},
  }
  \vspace{-0.4em}
  \begin{lstlisting}[]
      // b_i: index of the batch 
      // refsum_i: b_i-th reference summary of this batch
      // vocab: vocabulary size
      // seq_len: sequence length of reference summary
      // lm_logits: final output of the language model 

      softmax_prob = softmax(lm_logits, dim=-1)
      summary_prob = softmax_prob[b_i]
      summary_prob = summary_prob[:seq_len, :]
      prob_values, cand_indices = max(summary_prob, dim=-1)

      // 2-gram loss of b_i-th summary
      twoGram_loss_i = 1
      twoGrams = dict()

      // calculation of Equations (10) and (11)
      for s in range(seq_len - 1):
          k = [str(i) for i in refsum_i[s:s+2].tolist()]
          k = '_'.join(k)
          v = twoGrams.get(k)
          if v is None:
              v = []
          if all(cand_indices[s:s+2] == refsum_i[s:s+2]):
              v.append(prod(prob_values[s:s+2]))
          twoGrams[k] = v

      each_gram = 1 / (seq_len - 1)

      for k, v in twoGrams.items():
          num_grams = len(v)
          for prob in v:
              twoGram_loss_i -= (each_gram / num_grams * prob)
  
      // loop over every b_i summary
      // get the summation of twoGram_loss_i
      // eventually divided by batch size
  \end{lstlisting}
  \vspace{-0.5em}
\end{algorithm}

\begin{algorithm}[t]
  \caption{\small{2-gram matches in PyTorch-like style}}
  \label{alg:code match}
  \algcomment{\fontsize{7pt}{0em}\selectfont 
  \textrm{max}: torch.max(), return maximum values of the elements and their indices.\\
  \textrm{prod}: torch.prod(), return the product of all elements in the vetor.
  }
  \definecolor{codeBLEU}{rgb}{0.25,0.5,0.5}
  \lstset{
    backgroundcolor=\color{white},
      basicstyle=\fontsize{7pt}{7pt}\rmfamily\selectfont,
      columns=fullflexible,
      breaklines=true,
      captionpos=b,
      commentstyle=\ttfamily\selectfont\fontsize{8pt}{8pt}\color{black},
      keywordstyle=\color{black}\fontsize{7pt}{7pt},
      numbers=left,
      numbersep=-8pt,
      emph={
        softmax_prob,
        lm_logits
        summary_prob,
        prob_values,
        cand_indices,
        summary_prob,
        refsum_i,
        vocab,
        seq_len,
        lm_logits,
        twoGram_loss_i,
        twoGrams,
        s,
        k,
        i,
        v,
        each_gram,
        num_grams,
        prob,
        b_i
      },
      emphstyle={\itshape},
      morecomment=[l]{//},
  }
  \vspace{-0.4em}
  \begin{lstlisting}[]
      // b_i: index of the batch
      // refsum_i: b_i-th reference summary of this batch
      // vocab: vocabulary size
      // seq_len: sequence length of reference summary
      // lm_logits: final output of the language model

      softmax_prob = softmax(lm_logits, dim=-1)
      summary_prob = softmax_prob[b_i]
      summary_prob = summary_prob[:seq_len, :]
      prob_values, cand_indices = max(summary_prob, dim=-1)

      // 2-gram loss of b_i-th summary
      twoGram_loss_i = 1
      twoGrams = dict()

      // calculation of Equations (12) and (13)
      for s in range(seq_len - 1):
          k = [str(i) for i in refsum_i[s:s+2].tolist()]
          k = '_'.join(k)
          twoGrams[k] = []

      for s in range(seq_len - 1):
          k = [str(i) for i in cand_indices[s:s+2].tolist()]
          k = '_'.join(k)
          v = twoGrams.get(k)
          if v is not None:
              v.append(prod(prob_values[s:s+2]))
              twoGrams[k] = v

      each_gram = 1 / (seq_len - 1)

      for k, v in twoGrams.items():
          num_grams = len(v)
          for prob in v:
              twoGram_loss_i -= (each_gram / num_grams * prob)

      // loop over every b_i summary
      // get the summation of twoGram_loss_i
      // eventually divided by batch size
  \end{lstlisting}
  \vspace{-0.5em}
\end{algorithm}

\subsection{BART-base implementation}

We use BART-base \citep{Lewis2020}, an attention-based \citep{Vaswani2017} pre-trained language model with encoder-decoder architecture, including 6 layers of encoder, 6 decoder layers, 768 hidden states and 140M trainable parameters for implementation.
The model uses CE loss as training objective. Since we want to mitigate the discrepancy between the training criterion and the evaluating criterion, we jointly train the BART-base model with CE loss and $N$-gram objective. 
We manage to vary $N$-gram objectives and investigate which combination is the best.

Let $N_{max}$ indicate the $N$-gram objective  with respect to maximum $N$ that participates in the investigation of objectives, and the final target of $N$-gram rewards is:

\begin{equation}
  \mathcal{L} = \mathcal{L}_{CE} + \sum_{N=2}^{N_{max}} \mathcal{L}_{N \rm -gram\;rewards}(\theta)
  \label{eq:$N$-gram rewards final objective}
\end{equation}

Accordingly, the final target of $N$-gram matches can be computed as follows:
\begin{equation}
  \mathcal{L} = \mathcal{L}_{CE} + \sum_{N=1}^{N_{max}} \mathcal{L}_{N \rm -gram\;matches}(\theta)
  \label{eq:$N$-gram matches final objective}
\end{equation}

\section{Experiments}
\label{sec:exp}

In this section, we first introduce CNN/DailyMail dataset and XSum dataset. 
Second, we detail the training settings and inference settings of the language model.
Next, we describe the evaluation metrics as well as compared methods. 
Furthermore, we present experimental results with analysis and ablation study.
Finally, we show a few generated examples from the language model.

\subsection{Datasets}

We test the proposed algorithm by fine-tuning BART-base model on CNN/DailyMail dataset\footnote{\small https://huggingface.co/datasets/cnn\_dailymail} and XSum dataset\footnote{\small https://huggingface.co/datasets/xsum}, which are both commonly used news datasets for automatic abstractive summarization. 
CNN/DailyMail \citep{nallapati2016} contains news articles and the corresponding highlights from CNN and DailyMail, while XSum \citep{narayan2018} contains news articles from BBC with one-sentence summaries and has higher level of text compression. 
Statistics of the datasets are shown in Table~\ref{tab:stats_dataset}.

\begin{table}[t]
  \tablestyle{1pt}{1.3}
  \scalebox{0.9}{
  \begin{tabular}
  {x{36}|x{54}|x{26}x{26}|x{26}x{26}|x{52}} 
  \hline
  &
  \multicolumn{1}{c|}{} & \multicolumn{2}{c|}{Avg. Source} & \multicolumn{2}{c|}{Avg. Summary} & \multicolumn{1}{c}{$\%$ novel}
  \\
  Dataset & Train/Vaild/Test & Words & Sents & Words & Sents & bi-gram
  \\ \hline
  CNN/DM & 287K/13K/11K &
  690.90 & 42.38 & 49.08 & 3.81 & 62.22
  \\
  XSum & 204K/11K/11K &
  431.07 & 19.77 & 23.26 & 1.00 & 88.13
  \\ \hline
  \end{tabular}
  }
  \vspace{.5em}
  \caption{\textbf{Statistics of the summarization datasets.} 
  The average number of words and sentences for source and summary are counted before tokenization.
  $\%$ novel bi-gram represents the average proportion of new bi-grams that a reference summary contains but the corresponding source document does not.
  }
  \label{tab:stats_dataset}
  \vspace{0em}
\end{table}

\definecolor{Gray}{gray}{0.5}
\renewcommand{\hl}[1]{\textcolor{HighlightImprove}{#1}}
\newcommand{\resnormal}[2]{\tablestyle{1pt}{1} \begin{tabular}{z{16}y{25}} {#1} & {} \end{tabular}}
\newcommand{\resimprove}[3]{
  \tablestyle{1pt}{1} 
  \begin{tabular}{z{16}y{25}}
    {#1} &
    \fontsize{7.5pt}{1em}\selectfont{~\hl{(${#2}$#3)}}
  \end{tabular}
}

\newcommand{\resdecline}[3]{
\tablestyle{1pt}{1} 
\begin{tabular}{z{16}y{25}}
{#1} &
\fontsize{7.5pt}{1em}\selectfont{~\textcolor{HighlightDecline}{(${#2}$#3)}}
\end{tabular}}

\subsection{Experiment settings}
We implement the experiments on pre-trained BART-base model in PyTorch framework\footnote{\small https://huggingface.co/facebook/bart-base}. 
We add \texttt{[SEP]} token at the beginning of every sentence in the source documents during the preprocessing. 
Following the default hyperparameter settings from BART \citep{Lewis2020}, we apply 1024 as maximum source length and 128 as maximum target length for BART-base model. 
The model consists of 6 layers of encoder blocks,  6 layers of decoder blocks and 768 hidden states. 
We set batch-size of 16 and gradient accumulation steps as 2, therefore total training batch size is 32. 
We optimize the model with Adam optimizer ($\beta_{1} = 0.9$, $\beta_{2} = 0.999$, $\epsilon = 10^{-8}$), and  set weight decay as 0.01. 
We fine-tune the pre-trained model for 10 epochs and warm up the learning rate with 1000 steps from 0 to 3 $\times$ 10$^{-5}$, and the learning rate linearly declines afterwards. 
We evaluate the model on vailation dataset every 1000 training steps, eventually load and report the test result of model checkpoint with lowest evaluation loss on vailation dataset. 
We conduct the experiment on  2 RTX 3090 with mixed precision, which takes approximately 40 hours for training. 
For CNN/DM dataset, we set length penalty of 2, beam width of 4, maximum and minimum generation lengths of 142 and 56 respectively. 
For XSum dataset, we set length penalty of 1, beam width of 6, maximum and minimum generation lengths of 62 and 11 respectively.

\subsection{Evaluation Metrics}
We evaluate the summarization quality with the following metrics:
\begin{itemize}
  \setlength{\itemsep}{0pt}
  \item[1)]\textbf{ROUGE} measures overlapping units between  target summary and  machine generated summary. 
  We report F1-scores of ROUGE-1, ROUGE-2 and ROUGE-L of the experiment results. 
  The ROUGE scores are computed using \emph{rouge-score} package\footnote{\small https://pypi.org/project/rouge-score}.
  \item[2)]\textbf{Word Mover's Distance (WMD)} \citep{KusnerSKW15} quantizes the dissimilarity between two documents as the minimum cumulative distance that all tokens in a document move to another document's tokens in $word2vec$ embedded space \citep{word2vecMikolov2013}.
  WMD is computed with \emph{wmd} package\footnote{\small https://pypi.org/project/wmd}.
  \item[3)]\textbf{BERTScore} \citep{ZhangKWWA20} computes the similarity between a candidate text's contextualized embeddings and a reference text's contextualized embeddings. The embeddings are extracted from the pre-trained BERT model.
  BERTScore is computed with \emph{bert-score} package\footnote{\small https://pypi.org/project/bert-score}.
  \item[4)]\textbf{FactCCX} \citep{kryscinski2020factCC} 
    is a factual consistency check model with pre-trained BERT and can examine the sentence-level correlation between source document and claim document (\ie, candidate text).
    The model is fine-tuned on artificially augmented inconsistent samples.
  \item[5)]\textbf{SummaC\textsubscript{CONV}} \citep{LabanSBH22Summac}
    is a natural language inference (NLI) model (pre-trained BERT) for inconsistency detection. 
  The model computes the probabilities of every sentence over three properties (entailment, contradiction and neutral), bringing an NLI pair matrix across source document and  generated summary.
  The model passes the NLI pair matrix to a convolutional layer and computes the mean consistency scores of candidate documents.
\end{itemize}

\begin{figure*}
  \begin{minipage}[t][345pt]{1.0\textwidth}
  \begin{table}[H]
  \small
  \centering
  \tablestyle{1pt}{1.02}
  \subfloat[CNN/DailyMail\label{result on cnndm}]{
    \scalebox{0.87}{
    \begin{tabular}{p{2.8cm}|x{60}|x{60}|x{60}|x{60}|x{60}|x{60}|x{60}|x{60}}
      \hline  
      & R-1 ($\uparrow$) 
      & R-2 ($\uparrow$) 
      & R-L ($\uparrow$) 
      & BERTScore ($\uparrow$)
      & WMD ($\downarrow$)
      & FactCCX ($\uparrow$)
      & SummaC\textsubscript{CONV} ($\uparrow$)
      & Docs/s ($\uparrow$)
      \\ \hline 
    Lead-3 & \resnormal{40.34}{} & \resnormal{17.70}{} & \resnormal{36.57}{} & 87.08 & 4.5366 & \textbf{99.83} & \textbf{92.41} & - \\
    Pointer + covg. & \resnormal{39.53}{} & \resnormal{17.28}{} & \resnormal{36.38}{} & - & - & - & - & - \\
    BertSumExtAbs & \resnormal{42.13}{} & \resnormal{19.60}{} & \resnormal{39.18}{} & 85.30 & 4.6565 & ~5.60 & 26.47 & - \\
    T5 & \resnormal{43.52}{} & \resnormal{21.55}{} & \resnormal{40.69}{} & - & - & - & - & - \\
    ProphetNet & \resnormal{44.20}{} & \resnormal{21.17}{} & \resnormal{41.30}{} & 85.98 & 4.3665 & 13.04 & 92.35 & - \\
    GSum & \resnormal{45.95}{} & \resnormal{22.32}{} & \resnormal{42.48}{} & - & - & - & - & - \\
    SimCLS & \resnormal{46.67}{} & \resnormal{22.15}{} & \resnormal{\textbf{43.54}}{} & \textbf{88.44} & \textbf{4.0585} & 68.11 & 89.17 & - \\
    FES & \resnormal{\textbf{46.91}}{} & \resnormal{\textbf{22.84}}{} & \resnormal{43.47}{} & - & - & - & - & - \\
    BART-large & \resnormal{44.16}{} & \resnormal{21.28}{} & \resnormal{40.90}{} & 87.98 & 4.2653 & 80.83 & 87.44 & - \\
    \hline
    BART-base & \resnormal{43.26}{} & \resnormal{20.49}{} & \resnormal{40.40}{} & 88.23 & 4.3285 & 78.18 & 77.33 & \textbf{23.45 ($\times 1.00$)} \\
    + BoN & \resnormal{43.27}{} & \resimprove{20.60}{+}{0.11} & \resnormal{40.40}{} & 88.30 & 4.3369 & 77.14 & 78.02 & 21.48 ($\times 0.92$) \\
    + P-P2 & \resdecline{42.63}{-}{0.63} & \resdecline{20.30}{-}{0.19} & \resdecline{39.72}{-}{0.68} & 88.16 & 4.3917 & \textbf{78.89} & 76.89 & 19.91 ($\times 0.84$) \\
    + 2-gram rewards & \resimprove{43.46}{+}{0.20} & \resimprove{20.63}{+}{0.14} & \resimprove{40.62}{+}{0.22} & 88.29 & 4.3123 & 78.21 & 79.04 & 18.97 ($\times 0.81$) \\
    + 2-gram matches & \resimprove{43.33}{+}{0.07} & \resimprove{20.56}{+}{0.07} & \resimprove{40.56}{+}{0.16} & 88.28 & 4.3189 & 76.92 & 78.25 & 19.50 ($\times 0.83$) \\
    + (2,3,4)-gram rewards & \resimprove{\textbf{43.65}}{+}{0.39} & \resimprove{\textbf{20.81}}{+}{0.32} & \resimprove{40.88}{+}{0.48} & 88.34 & 4.2959 & 78.66 & \textbf{80.48} & 14.81 ($\times 0.63$) \\
    + (2,3,4)-gram matches & \resimprove{\textbf{43.65}}{+}{0.39} & \resimprove{\textbf{20.81}}{+}{0.32} & \resimprove{\textbf{40.90}}{+}{0.50} & \textbf{88.35} & \textbf{4.2951} & 78.36 & 80.21 & 16.53 ($\times 0.70$) \\
    \hline  
  \end{tabular}}}
  \vspace{-0.5em}

  \subfloat[XSum\label{result on xsum}]{
    \scalebox{0.87}{
    \begin{tabular}{p{2.8cm}|x{60}|x{60}|x{60}|x{60}|x{60}|x{60}|x{60}|x{60}}
      \hline
      & 
      R-1 ($\uparrow$) 
      & R-2 ($\uparrow$) 
      & R-L ($\uparrow$) 
      & BERTScore ($\uparrow$)
      & WMD ($\downarrow$)
      & FactCCX ($\uparrow$)
      & SummaC\textsubscript{CONV} ($\uparrow$)
      & Docs/s ($\uparrow$)
      \\ \hline 
    Lead-3 & \resnormal{16.30}{} & \resnormal{1.60}{} & \resnormal{11.95}{} & 85.61 & 6.1553 & \textbf{99.97} & \textbf{88.95} & - \\
    Pointer + covg. & \resnormal{28.10}{} & \resnormal{8.02}{} & \resnormal{21.72}{} & - & - & - & - & - \\
    BertSumExtAbs & \resnormal{38.81}{} & \resnormal{16.50}{} & \resnormal{31.27}{} & 87.31 & 5.0256 & 54.05 & 26.98 & - \\
    GSum & \resnormal{45.40}{} & \resnormal{21.89}{} & \resnormal{36.67}{} & - & - & - & - & - \\
    SimCLS & \resnormal{47.61}{} & \resnormal{24.57}{} & \resnormal{39.44}{} & \textbf{91.67} & \textbf{4.1744} & 48.52 & 25.02 & - \\
    FES & \resnormal{\textbf{47.77}}{} & \resnormal{\textbf{24.95}}{} & \resnormal{\textbf{39.66}}{} & - & - & - & - & - \\
    BART-large & \resnormal{45.14}{} & \resnormal{22.27}{} & \resnormal{37.25}{} & 91.62 & 4.1971 & 47.89 & 30.42 & - \\
    \hline
    BART-base & \resnormal{41.63}{} & \resnormal{18.75}{} & \resnormal{33.73}{} & 91.53 & 4.5206 & 48.76 & 24.33 & \textbf{29.00 ($\times 1.00$)} \\
    + BoN & \resimprove{41.90}{+}{0.27} & \resimprove{19.00}{+}{0.25} & \resimprove{34.12}{+}{0.39} & 91.56 & \textbf{4.5074} & \textbf{49.07} & 24.57 & 27.13 ($\times 0.94$) \\
    + P-P2 & \resdecline{40.95}{-}{0.68} & \resdecline{18.31}{-}{0.44} & \resdecline{33.03}{-}{0.70} & 91.44 & 4.5213 & 47.10 & 24.24 & 26.08 ($\times 0.90$) \\
    + 2-gram rewards & \resimprove{41.89}{+}{0.26} & \resimprove{18.97}{+}{0.22} & \resimprove{34.04}{+}{0.31} & 91.58 & 4.5147 & 48.56 & 24.49 & 25.41 ($\times 0.88$) \\
    + 2-gram matches & \resimprove{\textbf{41.93}}{+}{0.30} & \resimprove{19.00}{+}{0.25} & \resimprove{\textbf{34.18}}{+}{0.45} & \textbf{91.59} & 4.5228 & 48.52 & 24.55 & 26.23 ($\times 0.90$) \\
    + (2,3,4)-gram rewards & \resimprove{41.91}{+}{0.28} & \resimprove{\textbf{19.01}}{+}{0.26} & \resimprove{34.08}{+}{0.35} & 91.57 & 4.5205 & 48.74 & 24.57 & 22.14 ($\times 0.76$) \\
    + (2,3,4)-gram matches & \resimprove{41.77}{+}{0.14} & \resimprove{18.97}{+}{0.22} & \resimprove{34.03}{+}{0.30} & 91.56 & 4.5309 & 48.74 & \textbf{24.67} & 23.90 ($\times 0.82$) \\
    \hline  
  \end{tabular}}}
  \vspace{-0.8em}
  \caption{
    \textbf{Results on CNN/DM and XSum.} We evaluate the summarization outputs with ROUGE-1 (R-1), ROUGE-2 (R-1), ROUGE-L (R-L), 
    BERTScore, Word Mover's Distance (WMD), FactCCX, and SummaC\textsubscript{CONV}. 
    Throughput rates (Docs/s) during training are shown in the last column, and relative speed is shown inside the bracket with BART-base as the baseline. The best result is in bold face.
    }
    
  \label{tab:mainresult}
  \end{table}
  \end{minipage}
  \vspace{0em}
\end{figure*}

\subsection{Compared Methods and Parameters}
We take BART-base as our seq2seq baseline and collectively train the model with CE loss and $N$-gram rewards objective or $N$-gram matches objective. 
We compare the proposed algorithm with the following summarization approaches:
\begin{itemize}
  \setlength{\itemsep}{0pt}
  \item[1)]\textbf{Lead-3} \citep{See2017}: an extractive summarization baseline that takes the leading three sentences as the summarization.
  \item[2)]\textbf{Pointer-generator + coverage} \citep{See2017}: a standard seq2seq attentional model for abstractive summarization, improved with coverage mechanism that avoids token repetition.
  \item[3)]\textbf{BertSumExtAbs} \citep{Liu2019}: a two-stage fine-tuned summarization model based on BERT \citep{Devlin2019}, which is fine-tuned the encoder with extractive summarization dataset then fine-tuned the encoder-decoder model with abstractive summarization dataset.
  \item[4)]\textbf{T5} \citep{Raffel2019}: a self-supervised encoder-decoder model trained on large clean corpus, which views and converts all NLP tasks as seq2seq task.
  \item[5)]\textbf{BART} \citep{Lewis2020}: one of the best transformer-based self-supervised masked autoencoding model for seq2seq tasks. BART-large model has 12 layers of encoder, 12 layers of decoder, 1024 of hidden states, 50K corpus size and 406M of model parameters, while BART-base model has 6, 6, 768, 140M for encoder layers, decoder layers, hidden states and  model parameters respectively.
  \item[6)]\textbf{ProphetNet} \citep{Qi2020}: a self-supervised pre-trained model that optimizes n-step ahead token prediction. It is also a transformer-based language model, containing 12 layers of encoder, 12 layers of decoder, 1024 of hidden states, 30K corpus size and 380M of model parameters.
  \item[7)]\textbf{GSum} \citep{dou-etal-2021-gsum}: a transformer-based encoder-decoder summarization framework. Initializing the model weights with BART, the encoder was jointly trained with both the source document and external guidance signal, meanwhile an additional cross-attention block was introduced for the guidance signal. The guidance signal can be the abstractive or extractive summary generated by another fine-tuned summarization model.
  \item[8)]\textbf{SimCLS} \citep{liu2021simcls}: a two-stage abstractive summarization framework firstly requires a fine-tuned BART to generate multiple candidate summaries, and then trains a scoring model (RoBERTa) \citep{liu2019roberta} to rank the candidates through the reference summaries. Whilst the inference stage need both the candidates generation and candidates scoring.
  \item[9)]\textbf{FES} \citep{chen2022FES}: a faithfulness-enhanced summarization model with encoder-decoder architecture that integrates both the summarization and question-answering (QA) tasks. The encoder is optimized for the sequence representation of source document as well as answering faithfulness-related questions over the source document during the training process, while the decoder is for the summary generation. Importantly, The QA pairs were generated from a fine-tuned conditional QA generation model \citep{scialom2021}.
\end{itemize}

\subsection{Results}

In Table~\ref{tab:mainresult}, we report the evaluation results of our proposed method and compared methods whenever the fine-tuned checkpoints are available.

As shown in Table~\ref{result on cnndm}, we present the result of adding $N$-gram rewards objective on the BART-base model and fine-tuning it on the CNN/DM dataset.
Increase of 0.39/0.32/0.48 in ROUGE scores at R-1/R-2/R-L is achieved by (2,3,4)-gram rewards, respectively;
increase of 0.39/0.32/0.50 at R-1/R-2/R-L is respectively achieved by (2,3,4)-gram matches, where the R-L score even reaches BART-large's level: 40.90. It can be found that BoN brings about a subtle improvement on R-2: +0.11, while the overall ROUGE scores are dropped with P-P2. Also, we can find that 
improvements of 0.20/0.14/0.22 at R-1/R-2/R-L are achieved with 2-gram rewards objective, while 2-gram matches objective just marginally impact the ROUGE scores.

Table~\ref{result on xsum} provides the results on XSum. 
Since XSum dataset has higher requirement on abstractive capabilities of a language model, 
the overall evaluation scores are lower than CNN/DM's. In this setting, simply optimizing 2-gram works the best (increase of 0.30/0.25/0.45 at R-1/R-2/R-L is achieved respectively), whilst longer sub-sequence optimization may start overfitting training examples.
Accumulating other $N$-gram objectives achieve limited improvement, this could be because XSum examples have shorter word length and more novel bi-grams.
For XSum, rewarding too many probabilistic $N$-grams may rather harshly shape the probability distribution of the language model, therefore decreasing the performance. Note that P-P2 is still declining the ROUGE scores and BoN provides the second best improvement on R-1/R-2/R-L: 0.27/0.25/0.39, indicating that highly abstractive summaries with fewer target tokens (\ie, short summary length) can be preferable for BoN matching.


The evaluation of BERTScore between the reference summary and generated summary shows that our proposed method is slightly better than the competitors on CNN/DM and XSum.
Moreover, the result of ROUGE scores and BERTScore are generally positively correlated. 
BERTScore represents the cosine similarity between two document's high-dimensional embeddings from pre-trained BERT, 
hence it can be debatable that at what BERTScore level does a summary reach the human-level criterion for text summarization is not intuitive.
Our proposed method achieves the best BERTScore, which may be closer to human-level criteria. 

The evaluation results of WMD metric indicate that our proposed objectives achieve a lower dissimilarity level than the competing objective on CNN/DM.
However the evaluation on XSUM offers opposite evidences that there is no significant similarity difference between the original BART-base and the BART-base with our objectives. 
This could be because WMD metric may be less tolerant to alternative sub-sequence or sentence rephrasing.
Additionally, word embeddings that WMD metric applied might be less effective than BERTScore, which is self-supervised pre-trained on large corpura.

In terms of factual consistency evaluation, the FactCCX model may perceive marginal contextual consistency differences among the methods on the XSum dataset, 
\ie BoN objective holds a subtle advantage, while P-P2 hold a relative low score.
Nevertheless, on the CNN/DM dataset, the rankings of P-P2 and BoN are upside down, which is contradictory to the evaluation of ROUGE scores and BERTScore.
This might be due to FactCCX model is fine-tuned with artificially augmented inconsistent statement, thus the model may recall some false negative sub-sequences in the generated summaries and fail to technically indicate the statement consistency.

Next, the result of SummaC\textsubscript{CONV} shows that our proposed objectives achieve greater factual consistency than the competitors.
For the CNN/DM dataset, a positive correlation between the SummaC\textsubscript{CONV} score, ROUGE score and BERTScore can be observed.
All of our proposed objectives gain less inconsistency than the competing methods.
Generally, the consistency scores on CNN/DM are significantly higher than those on XSum.
The XSum dataset has only one sentence for each summary, therefore the factual consistency detection model could recall many false negatives of sub-sequences from the source document and hardly be effective for evaluating extremely abstractive summary.

Lead-3 reaches the highest factual consistency scores because it simply copies the first three sentences from the source document as the summary, which is unlikely against the claims in the source document.
Further, with deeper network layers and larger embedding spaces, BART-large achieves overall better scores than BART-base. Moreover, SimCLS accomplishes top-tier contextual similarities over models that have fewer supervised training stages such as BART and ProphetNet. 
However, SimCLS generates candidates through fine-tuned BART-large and then selects the best summary with a trained scoring model, 
yet the system does not promisingly win over BART-large on factual consistency metrics, 
which reveals there might be a trade-off between factual consistency metrics and other metrics.

The results show that training with our proposed objectives would spend more time than the original BART model with/without BoN objective.
However, the current vanilla implementation can be accelerated by storing the reference summary's $N$-gram during preprocessing to achieve higher throughput, especially for accumulating $N$-gram objectives.
We remark that further code optimization and code decoupling can effectively shorten the training time. 
Further, it is noted that the time consumption of $N$-gram rewards objective is greater than thoes of others.
The reason is slicing tensor objects and comparing tensors with $torch.all()$ function is usually time-consuming,
which takes no more than 60\% running time while achieves 4\% performance improvement in SummaC\textsubscript{CONV} (as shown in Table \ref{result on cnndm}). 
Thus, considering the complexity reward trade-off, our proposal has potential for applications.

GSum, SimCLS and FES were implemented with two fine-tuned language model, 
specifically BART-large is a favourable pre-trained backbone network in these state-of-the-art (SOTA) systems. 
It can be found that an increase of roughly 2.5 on R-1 is achieved with SimCLS and FES on both CNN/DM and XSum. 
However, the SOTA systems require auxiliary and contextual signals through additional fine-tuned model that can considerably increase the memory usage and computational complexity. 
Our proposed method reached practical improvements over BART-base, which can be regarded as an enrichment of operational use for BART.

We have released all generated examples\footnote{\small github.com/zhuyunqi96/ngramObj/tree/main/predictexample} of the test dataset of CNN/DM and XSum in Table~\ref{result on cnndm} and Table~\ref{result on xsum}.
In particular, we present output examples and underline their overlapping sub-sequences over reference summary, generated summaries and the corresponding document in Section \ref{sec:outputexamples} for case study.

\subsection{Ablation analysis}
\label{sec:ablation}

We list ablation results on CNN/DM dataset with $N$-gram rewards in Table~\ref{ablation rewards} and $N$-gram matches in Table~\ref{ablation matches}, where
the first row illustrates the result of BART-base model with only original CE loss.


\begin{table}[t]
  \centering
  \footnotesize
  \subfloat[$N$-gram rewards \label{ablation rewards}]{
  \scalebox{0.915}{
  \begin{tabular}{p{1.9cm}|x{28}|x{28}|x{28}}
  \hline
   & R-1 & R-2 & R-L
  \\ \hline
  BART-base & 43.26 & 20.49 & 40.40
  \\
  + 2-gram & \textcolor{greyColor}{\textbf{43.46}} & 20.63 & \textcolor{greyColor}{\textbf{40.62}}
  \\
  + 3-gram & 43.38 & 20.60 & 40.57
  \\
  + 4-gram & 43.40 & \textcolor{greyColor}{\textbf{20.69}} & 40.61
  \\
  + 5-gram & 43.26 & 20.47 & 40.45
  \\ \hline
  + 2,3-gram & 43.63 & 20.76 & 40.84
  \\
  + 2,3,4-gram & \textbf{43.65} & \textbf{20.81} & \textbf{40.88}
  \\
  + 2,3,4,5-gram & 43.43 & 20.64 & 40.65
  \\ \hline
  \end{tabular}
  }
  }
  \vspace{-0.5em}
  \footnotesize
  \subfloat[$N$-gram matches \label{ablation matches}]{
  \scalebox{0.915}{
  \begin{tabular}{p{1.9cm}|x{28}|x{28}|x{28}}
  \hline
   & R-1 & R-2 & R-L \\ 
  \hline
  BART-base & 43.26 & 20.49 & 40.40
  \\
  + 1-gram & 43.24 & 20.52 & 40.44
  \\
  + 2-gram & 43.33 & 20.56  & \textcolor{greyColor}{\textbf{40.56}}
  \\
  + 3-gram & 43.34 & 20.52 & 40.50
  \\
  + 4-gram & \textcolor{greyColor}{\textbf{43.36}} & \textcolor{greyColor}{\textbf{20.58}} & 40.52
  \\
  + 5-gram & 43.34 & 20.55 & 40.54
  \\ \hline
  + 1,2-gram & 43.45 & 20.64 & 40.63
  \\
  + 1,2,3-gram & 43.54 & 20.76 & 40.79
  \\
  + 1,2,3,4-gram & \textcolor{greyColor}{\textbf{{43.58}}} & \textcolor{greyColor}{\textbf{20.78}} & \textcolor{greyColor}{\textbf{40.85}}
  \\
  + 1,2,3,4,5-gram & 43.50 & 20.73 & 40.77
  \\ \hline
  + 2,3-gram & 43.29 & 20.58  & 40.52
  \\
  + 2,3,4-gram & \textbf{43.65} & \textbf{20.81} & \textbf{40.90}
  \\
  + 2,3,4,5-gram & 43.53 & 20.71 & 40.76
  \\ \hline
  \end{tabular}
  }
  }
  \vspace{-0.5em}
  \caption{\textbf{Ablation of $N$-gram rewards objective and $N$-gram matches objective on CNN/DM.} 
  The experiments are implemented on BART-base model, we jointly optimize the objective with cross-entropy loss.
  The best result is in bold face.
  }
  \label{tab:ablation table}
  \vspace{-1.0em}
\end{table}

\paragraph{$N$-gram rewards.}
It can be noticed that the accumulation of 2-gram, 3-gram and 4-gram rewards produces better improvement over baseline, while 5-gram rewards can damage the performance.
If only utilizing one $N$-gram rewards, 2-gram raises R-1/R-2/R-L: 0.20/0.14/0.22, outperforming the other $N$-gram rewards objectives. 
The effectiveness of 2-gram, 3-gram and 4-gram rewards may be due to they are rewarding solid and semantically coherent sub-sequences, and adequately guide the parameters of language model for generating proper probabilities distribution.
For 5-gram rewards, when it is introduced solely, too few 5-gram matches are spotted to achieve satisfying effectiveness.
When 5-gram rewards is jointly optimized with other $N$-gram rewards, the negative effect may due to the relatively long sub-sequences which can make the model overfit the summary sample during training.

\paragraph{$N$-gram matches.}
When only one $N$-gram matches objective is jointly optimized with CE loss, ROUGE scores can be enhanced with the objectives, but not as much as $N$-gram rewards. 
Besides, 5-gram matches provide moderate improvement, whereas 5-gram rewards bring unnoticeable changes. 
Since unigram matches objective produces subtle differences, we further investigate the performance of $(2,...,N)$-gram matches.

Stacks of $N$-gram matches objectives can offer improvement until 5-gram matches objective is involved.
The phenomenon is consistent with the experiment results of $N$-gram rewards.
Although the involvement of 5-gram objective continue suppressing the overall performance, the objectives of (1,2,3,4,5)-gram matches and (2,3,4,5)-gram matches outperform the objective of (2,3,4,5)-gram rewards.
Furthermore, (2,3,4)-gram matches objective offers almost the same enhancement with (2,3,4)-gram rewards.
This could be because $N$-gram matches objectives can have less repeated optimization on compounded sub-sequences, whereas accumulating $N$-gram rewards objectives cannot avoid over-rewarding sub-sequences.


\subsection{Case study}
\label{sec:outputexamples}

In this section, we present some generated examples from CNN/DM (Table~\ref{tab:example_1}) and XSum (Table~\ref{tab:example_2} and Table~\ref{tab:example_3}).

Table~\ref{tab:example_1} shows that our proposed objectives (except 2-gram matches objective) successfully recognize the salient sub-sequence: "tipped to replace Steve McClaren" for subject "Scan Dyche".
As all of the objectives fail to highlight "Dyche ... cannot comprehend", it indicates that recognization and compression of high-level context may be still dependent on the capability of a language model.
Table~\ref{tab:example_2} shows that our objectives of 2-gram rewards and 2-gram matches objectives can weight sub-sequence "special discussion" to be more important than "parallel talks".
Although the output of original BART-base contains "special discussion", the condition in its summary "over its future relationship with the EU" is not as precise as the others.
Table~\ref{tab:example_3} presents a challenging case that the model badly compresses and rephrases the context, only the (2,3,4)-gram matches objective catches one keyword "mental health".

Overall, the above examples demonstrate that more sub-sequence matches could be achieved with the proposed method.
In fact, our $N$-gram objectives aim to improve the probabilistic distribution of sequence generation at sub-sequence level,
hence the current objectives cannot correctly preceive factual context like "special discussion" and "parallel talks" in Table~\ref{tab:example_2}.
A comprehensive fact-aware summarization model with $N$-gram objectives is worthy of further in-depth study.

\ULforem
{
\begin{table}[H]
\centering
\tablestyle{1pt}{1.0}
\begin{footnotesize}
\scalebox{0.915}{
\begin{tabular}{p{0.5\textwidth}}
\hline
\begin{small}\textbf{Document: }\end{small} 
$[...]$
Burnley manager Sean Dyche, aka The Ginger Mourinho, \uwave{insists he is enjoying life at Turf Moor.} 
$[...]$
Dyche has been \uwave{tipped to replace former England manager Steve McClaren (right) at Derby.} 
$[...]$
\uwave{Some Arsenal fans called for Arsene Wenger's} head following a difficult start to the season. 
Dyche remains a big fan of the Gunners boss and cannot comprehend why there were calls for the three-time Premier League winner to be sacked. 
$[...]$
\\ \hline
\begin{small}\textbf{Reference: }\end{small} 
Sean Dyche has been \uwave{tipped to replace Steve McClaren at Derby.} 
Burnley boss insists he is \uwave{enjoying the 'challenge' at Turf Moor.} 
Dyche can't understand why Arsene Wenger gets stick from Arsenal fans.
\\ \hline
\begin{small}\textbf{BART-base: }\end{small} 
Sean Dyche has been linked with a move to Derby County.
The Clarets boss \uwave{insists he is enjoying life at Turf Moor.} 
Dyche has retained the support of the Burnley fans throughout the season.
\\ \hline
\begin{small}\textbf{BART-base + BoN: }\end{small} 
Sean Dyche has been linked with a move to Derby County.
The Clarets boss \uwave{insists he is enjoying life at Turf Moor.} 
Dyche has retained the support of the Burnley fans throughout the season.
\\ \hline
\begin{small}\textbf{BART-base + PP2: }\end{small} 
Sean Dyche \uwave{insists he is enjoying life at Turf Moor.} 
The Burnley boss has been \uwave{tipped to replace Steve McClaren at Derby.} 
Dyche says he is not looking further afield at the Clarets.
\\ \hline
\begin{small}\textbf{BART-base + 2-gram rewards: }\end{small} 
Sean Dyche has been \uwave{tipped to replace Steve McClaren at Derby.} 
The Clarets are currently second from bottom in the Premier League.
Dyche \uwave{insists he is enjoying life at Turf Moor.}
\\ \hline
\begin{small}\textbf{BART-base + 2-gram matches: }\end{small} 
Sean Dyche has been linked with a move to Derby County.
The Clarets boss \uwave{insists he is enjoying life at Turf Moor.} 
Dyche is also keen on replacing Steve McClaren at Newcastle.
\\ \hline
\begin{small}\textbf{BART-base + (2,3,4)-gram rewards: }\end{small} 
Sean Dyche has been \uwave{tipped to replace Steve McClaren at Derby.} 
The Burnley boss \uwave{insists he is enjoying life at Turf Moor.} 
Dyche has retained the support of the Burnley fans throughout the season.
Some Arsenal fans called for Arsene Wenger to be sacked.
\\ \hline
\begin{small}\textbf{BART-base + (2,3,4)-gram matches: }\end{small} 
Sean Dyche has been \uwave{tipped to replace Steve McClaren at Derby.} 
The Clarets are currently second from bottom in the Premier League.
Dyche \uwave{insists he is enjoying life at Turf Moor.} 
\\ \hline
\end{tabular}
}
\end{footnotesize}
\vspace{-1em}
\caption{generated example on CNN/DM \#1856.}
\label{tab:example_1}
\end{table}
}

{
\begin{table}[H]
\centering
\tablestyle{1pt}{1.0}
\begin{footnotesize}
\scalebox{0.915}{
\begin{tabular}{p{0.5\textwidth}}
\hline
\begin{small}\textbf{Document: }\end{small} 
Official discussions on the terms of the UK's exit and its future relationship with the EU are expected to begin in the Spring once the UK has triggered Article 50 - notifying the union of its intention to leave.
$[...]$
Speaking after holding talks with Austrian Chancellor Christian Kern, Mr Juncker said there would be no parallel negotiations and \uwave{the UK would not be allowed} to conduct bilateral discussions in \uwave{key areas} such as finance, telecommunications or chemicals.
$[...]$
"It is in our interests therefore that \uwave{we don't have any special discussions}... with certain individual countries."
$[...]$
\\ \hline
\begin{small}\textbf{Reference: }\end{small} 
The UK should not try to play different EU states off against each other or \uwave{pursue "special discussions" in key areas}, a top EU official has warned.
\\ \hline
\begin{small}\textbf{BART-base: }\end{small} 
The European Union has said it will not have any \uwave{special discussions} with the UK over its future relationship with the EU.
\\ \hline
\begin{small}\textbf{BART-base + BoN: }\end{small} 
The European Union has said there will be no parallel negotiations on the UK's exit from the bloc.
\\ \hline
\begin{small}\textbf{BART-base + PP2:}\end{small} 
The European Union has said there will be no parallel talks between the UK and the rest of the bloc as it prepares to leave.
\\ \hline
\begin{small}\textbf{BART-base + 2-gram rewards:}\end{small} 
The EU has said there will be no \uwave{special talks} with the UK over its exit from the bloc.
\\ \hline
\begin{small}\textbf{BART-base + 2-gram matches:}\end{small} 
The UK will not be allowed to have \uwave{"special discussions"} with other EU countries on the terms of its exit from the bloc, the EU has said.
\\ \hline
\begin{small}\textbf{BART-base + (2,3,4)-gram rewards:}\end{small} 
The EU has said there will be no parallel talks with the UK over its exit from the bloc.
\\ \hline
\begin{small}\textbf{BART-base + (2,3,4)-gram matches:}\end{small} 
The UK will not be allowed to conduct bilateral discussions with certain countries in \uwave{key} economic zones and certain sectors, the EU has said.
\\ \hline
\end{tabular}
}
\end{footnotesize}
\vspace{-1em}
\caption{generated example on XSum \#630.}
\label{tab:example_2}
\end{table}
}

{
\begin{table}[H]
\centering
\tablestyle{1pt}{1.0}
\begin{footnotesize}
\scalebox{0.915}{
\begin{tabular}{p{0.5\textwidth}}
\hline
\begin{small}\textbf{Document: }\end{small} 
$[...]$
It also applauded steps to expand community supervision programmes, saying a rapid increase in the number of prisoners who are \uwave{seriously mentally ill} was putting the system under strain.
But it said "serious systemic deficiencies" and "deplorable environmental conditions" remained for those being held at the jail, \uwave{violating the US constitutional ban on cruel and unusual punishment.}
$[...]$
but the prison system throughout the state of California, where inmates serve longer sentences, has also come under scrutiny for overcrowding.
\\ \hline
\begin{small}\textbf{Reference: }\end{small} 
The condition of \uwave{mental healthcare} inside Los Angeles County jails is so poor that it is \uwave{unconstitutional}, the US justice department has said.
\\ \hline
\begin{small}\textbf{BART-base: }\end{small} 
California's jails have been criticized by the US justice department for overcrowding and overcrowding.
\\ \hline
\begin{small}\textbf{BART-base + BoN: }\end{small} 
California's San Bernardino County jail has been accused of "serious systemic deficiencies" in its handling of inmates.
\\ \hline
\begin{small}\textbf{BART-base + PP2: }\end{small} 
San Bernardino County Jail in California has "serious systemic deficiencies" that are putting the jail under strain, according to a US justice department report.
\\ \hline
\begin{small}\textbf{BART-base + 2-gram rewards: }\end{small} 
California's jails have been criticized by the US justice department for overcrowding and overcrowding.
\\ \hline
\begin{small}\textbf{BART-base + 2-gram matches: }\end{small} 
The US justice department has said it has found "serious systemic deficiencies" in the prison system in California.
\\ \hline
\begin{small}\textbf{BART-base + (2,3,4)-gram rewards: }\end{small} 
California's jails have been criticized by the US justice department for "serious systemic deficiencies" in their handling of \uwave{mental health issues.}
\\ \hline
\begin{small}\textbf{BART-base + (2,3,4)-gram matches:}\end{small} 
California's jail system has "serious systemic deficiencies" and "significantly understates" efforts to improve inmate safety, the US justice department has said.
\\ \hline
\end{tabular}
}
\end{footnotesize}
\vspace{-1em}
\caption{generated example on XSum \#3636.}
\label{tab:example_3}
\end{table}
}

\normalem


\section{Limitation and Conclusion}

The main limitation of our proposed method is that the number of $N$-gram objectives shall be involved can be a hyperparameter for different datasets.
Additionally, BART was pre-trained on Wikipedia and books corpus, and we only investigate our method on news summarization datasets which have moderate text length.
Applying the objectives on different pre-trained language models and conducting test on different categories of summarization datasets need further study.

In this paper, we propose two differentiable $N$-gram objectives based on probabilistic sub-sequence matching that no longer ceil the matched number of sub-sequences and can value the matched probabilistic sub-sequences equally and produce fairly significant improvement on abstractive summarization. The proposed objectives outperform P-P2 objective and BoN objective.
Meanwhile, applying our proposed algorithm on other seq2seq tasks is feasible.




\section{Acknowledgement}
This research was supported by the National Key R\&D Program of
China (No. 2021ZD0111000), the National Natural Science Foundation
of China (Nos. 61961160707 and 61976212), and Hainan Provincial Natural Science Foundation of China (No. 622RC618).

\bibliography{arxiv-ver}


\end{document}